\documentclass[letterpaper, 10 pt, conference]{ieeeconf}  

\IEEEoverridecommandlockouts                              

\usepackage{graphicx} 
\usepackage{mathptmx} 
\usepackage{amsmath} 
\usepackage{amssymb}  
\usepackage{subfig}
\usepackage[table,xcdraw]{xcolor}

\title{\LARGE \bf
Collaborative Localization of Aerial and Ground Mobile Robots through Orthomosaic Map
}

\author{Xuecheng Xu$^{1,*}$, Zexi Chen$^{1,*}$, Jiaxin Guo$^{1,*}$, Yue Wang$^{1}$, Yunkai Wang$^{1}$ and Rong Xiong$^{1,\dagger}$
\thanks{$^*$ Contributed equally to this manuscript}
\thanks{$^{1}$ Xuecheng Xu, Zexi Chen, Yue Wang, Yunkai Wang and Rong Xiong are with the State Key Laboratory of Industrial Control Technology and Institute of Cyber-Systems and Control, Zhejiang University, Zhejiang, China}%
\thanks{$^\dagger$ Corresponding author
\hspace{1.5ex} Email Address: {\tt\small rxiong@zju.edu.cn}}%
}
\begin{document}
\maketitle
\begin{abstract}

With the deepening of research on the SLAM system, the possibility of cooperative SLAM with multi-robots has been proposed. This paper presents a map matching and localization approach considering the cooperative SLAM of an aerial-ground system. The proposed approach aims to help precisely matching the map constructed by two independent systems who have large scale variance of view points of the same route and eventually enables the ground mobile robot to localize itself in the global map given by the drone. It contains dense mapping with Elevation Map and software ``Metashape", map matching with a proposed template matching algorithm, weighted normalized cross correlation (WNCC) and localization with particle filter. The approach enables map matching for cooperative SLAM with a feasibility of multiple scene sensors, varies from stereo cameras to lidars, and is insensitive to the synchronization of the two systems. We demonstrate the accuracy, robustness, and the speed of the approach under experiments of the Aero-Ground Dataset\cite{c1}.

\end{abstract}

\section{INTRODUCTION AND REVIEW}

With wider applications of mobile robots, the SLAM system has been a heated topic and has been used in a vast variety of applications in the past few years. The ability of localizing the robot and mapping its sensed environment under the vacancy of global positioning systems has been treated as one of the key features of the SLAM system. With the increasing demand for mobile robots, the working environment of robots has become more and more diverse. Some places that are lack of global positions have relatively level grounds and simple feasible regions could be easy for robots to pass through, such as indoors. However, there will inevitably be some non-position places to explore that are human-unreachable, obstructive, and even dangerous for robot to pass through.

\begin{figure}[t]
\centering
\includegraphics[scale=0.1]{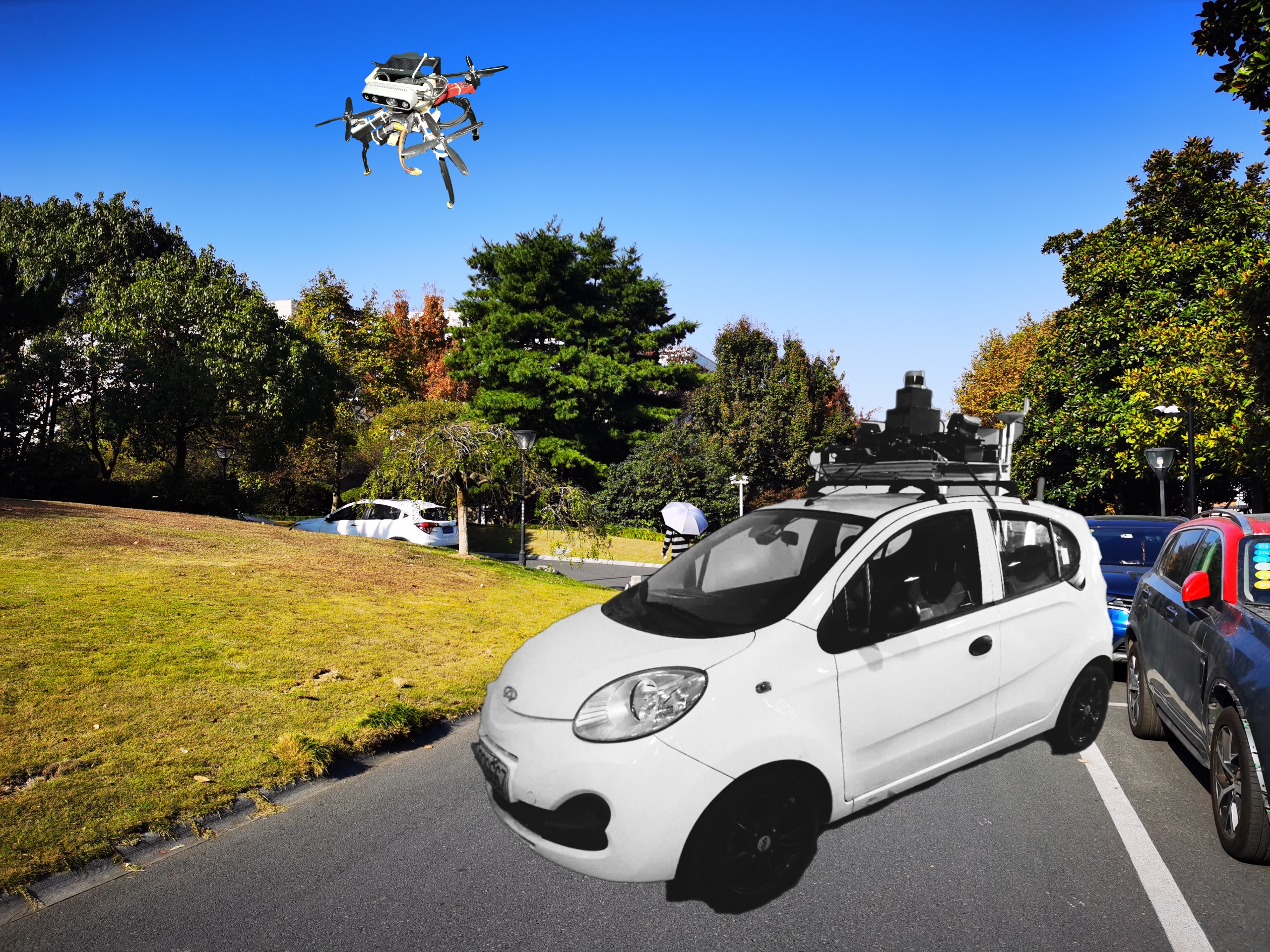}
\caption{Demonstration of collaborative SLAM carried by a ground mobile robot and a MAV with a stereo camera on both of them.}
\label{fig1} 
\end{figure}

In the mean time, with the rising of the micro-aerial vehicles (MAVs), the feasible region of one SLAM system has been greatly extended. Unknown places like obstructive caves could be roughly explored by light-weight MAVs and then detailed inspected by heavy-loaded ground mobile robots with multiple equipments following the map given by the MAV, as shown in Fig. \ref{fig1} . To ensure the collaboration of the two systems, an assisting mapping and localization is required.

Most of the aerial-ground collaboration focuses on the perception of a single robot and guides another one to achieve a particular goal. For example, a MAV can use the QR code\cite{c2} and other visual features to identify and track the ground mobile robot. There are also scholars using the camera mounted on the ground mobile robot to track the LED light of the MAV\cite{c3}. 

In addition, the aerial-ground collaboration framework proposed in \cite{c4} uses the visual features in the common perspective of an aerial-ground system to complete the aerial-ground map matching and localization. The disadvantage of this framework is that the visual features used for map matching such as points and lines will gradually become non-uniform as the viewing angle difference between the two perspective gradually becomes larger. For example, a line feature in the ground perspective is likely to become a point from its bird's-eye view. Therefore, this framework is limited by the scale of variance of the perspective.

The aerial-ground collaboration method proposed by Michael et al. \cite{c5} uses the rough relative position between the robots as an initial guess, and match the two maps within a range from the initial guess using iterative closest point algorithm (ICP). The result shows that it will give a precise match of the two map but with a large cost of MAV carrying a heavy-loaded lidar to complete the the mapping. The endurance of an MAV under heavy-loaded conditions has become a major obstacle to the success of this method.

Käslin et al. \cite{c6} proposed an aerial-ground collaboration method based on Elevation Map \cite{c7}. It uses depth information to establish Elevation Maps from both robots and completes aerial-ground matching by comparing the similarities of obstacles' heights within them by template matching. By using accurate depth information, this method can accurately and swiftly complete map matching and robot localization. However, without lidar equipped, depth information becomes obscure when the whole system works outside or when the MAV flies higher. Therefore, the application of this framework is limited by the working environment.

With the inspiration of the work \cite{c6}, template matching is adopted in our work. Traditional template matching approaches such Sum of Square Difference(SSD)\cite{c8}, Sum of Absolute Difference(SAD)\cite{c9}, and Normalized Cross Correlation(NCC)\cite{c10} share the same thought of sliding window with their own pros and cons.

The map matching and robot localizing system proposed in this paper aims at ensuring the collaboration with overcoming the poor depth conditions that an outdoor environment provide. The presented method constructs a colorized local Elevation Map for the ground mobile robot by endowing RGB data of every pixel acquired from the stereo camera to the 2.5D Elevation Map proposed by \cite{c12} and eventually forms the identical colorized orthomosaic map of every local frame. With colorized orthomosaic maps constructed for both the ground mobile robot's local map and the MAV's global map, the system is able to match both maps and gives the local map's real-time relative location to the global one by applying a particle filter. 

Due to the different approaches used by the aerial-ground mobile robots for mapping, the quality and style of the local and global orthomosaic maps varies greatly from each other. Therefore, the simple application of template matching based on sum differences is unsuitable. To deal with this, a new template matching, Weighted Normalized Cross Correlation(WNCC), is proposed.

This method does not require any expensive or heavy equipment from the short-lasting MAV and is insensitive to the depth information given by the MAV. Therefore, the MAV is able to form a more valuable global map with this method by flying further and higher.

The remainder of paper is organized as follows. The system constructs orthomosaic maps for the two robots with different approaches introduced in Section II. Every local orthomosaic map is matched with the global one with the WNCC algorithm introduced in Section III. In Section IV, with the matching transformation given of every local frame, the ground mobile robot localizes itself in the global map by applying a particle filter. The evaluation of our approach is shown in Section V and We conclude the paper and discuss future works in Section VI.

\section{MAPPING}

With wide vision and outstanding mobility, the MAV is able to construct a dense global map, and provide it as a proiri map for the ground mobile robot. Considering the inaccuracy of the online estimated depth when the MAV is flying up high outdoor, our method uses the 3D modeling software ``Metashape"\cite{c11} to construct the aerial orthomosaic map online with a reasonable time of delay for data transferring.

Ground mobile robots can carry lasers, stereo cameras, etc. in outdoor scenes to build local maps. In view of the navigation planning requirements of mobile robots, our method uses a real-time 2.5D Elevation Map with texture information\cite{c12} to construct a local map of the ground, and generates an orthomosaic map based on the 2.5D map rendering, which can be matched with the orthomosaic global map constructed by the MAV in the air for localization.

\subsection{Construction of Global Orthomosaic Map} 

\begin{figure}[h]
\centering
\includegraphics[scale=0.65]{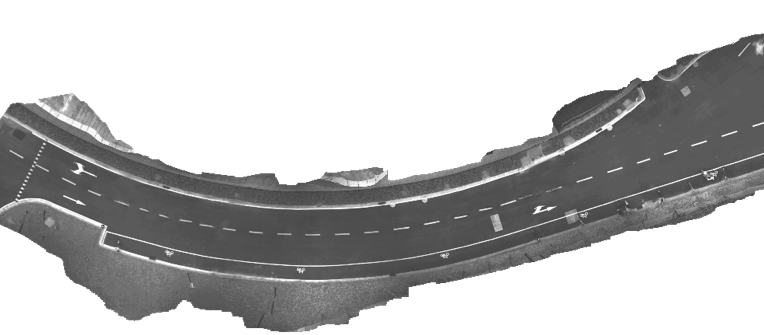}
\caption{Global orthomosaic map constructed by MAV with detailed and accurate texture. }
\label{fig2} 
\end{figure}

Our approach uses the commercial software  ``Metashape" (30-day free trial) to construct an aerial orthographic map. The MAV takes photos of the target area through the bird's-eye camera to obtain a photo sequence with timestamp stamped. By combining the timestamped-photos sequence of the bird's-eye camera with its intrinsic and extrinsic parameters, the software is able to estimate the camera pose and reconstruct a three-dimensional model of the contents in the photos. After rendering an orthomosaic map of the target area, which is the global map in this work, estimating the corresponding scaling information, we finally generate an orthomosaic global map with an absolute scale of 10 Centimeters corresponds to 1 pixel, shown in Fig. \ref{fig2}.

\subsection{Construction of Local Orthomosaic Map}

Considering the real-time performance, the computational cost, and the sparse 3D information required by the ground mobile robot, our approach adopted and upgraded the 2.5D Elevation with RGB data stored in every grid to build local orthomosaic maps. Moreover, Elevation Map's map representation is very suitable for mobile robot's navigation due to its characteristic of storing the height information of the corresponding area.

\begin{figure}[t]
\centering
\includegraphics[scale=0.3]{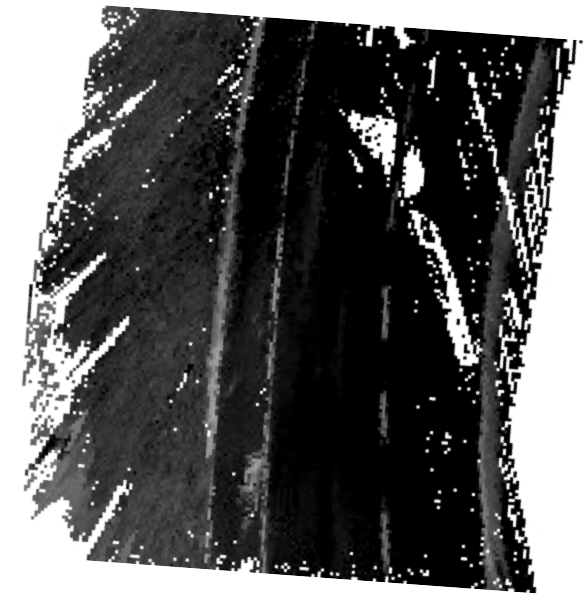}
\caption{Local orthomosaic map constructed by ground mobile robot with noisy background. }
\label{fig3} 
\end{figure}

The ground mobile robot calculates the depth with only a stereo camera and feed the depth information to the Elevation Map to build a 2.5D map. In the mean time, by feeding the RGB data and texture information to every grid of the map, we are able to construct a colorized intuitive local 2.5D map, abbreviated as local grid map. The local grid map also has absolute scale, and the size of each grid can be specified manually. Finally, as shown in Fig. \ref{fig3}, the approach renders the local grid map to the ground plane to generate a local orthomosaic map with an absolute scale of 10 cm for 1 pixel.

\section{MAP MATCHING}

Due to the different mapping approaches used for constructing global and local orthomosaic map, where ``Metashape" for global(Fig. \ref{fig2}) and Y. Pan's Elevation Map for local(Fig. \ref{fig3}), the map accuracy, map style and even features shown in both maps suffer drastic deviation. The global map constructed by ``Metashape" has good details and low error controlled within $\pm$0.1 meters when the local map constructed with a inaccurate calculated depth from a stereo camera suffering from its own hollowing, noise and error that can not be ignored. 

The huge difference in the quality of two maps has brought some difficulties to map matching. In order to achieve map matching with difficulties mentioned above, we proposes a weighted normalized product correlation algorithm, abbreviated as WNCC, through experiments, comparisons and calculations.

\subsection{Failure of SSD and SAD}

\begin{figure}[t]
\centering
\includegraphics[scale=0.3]{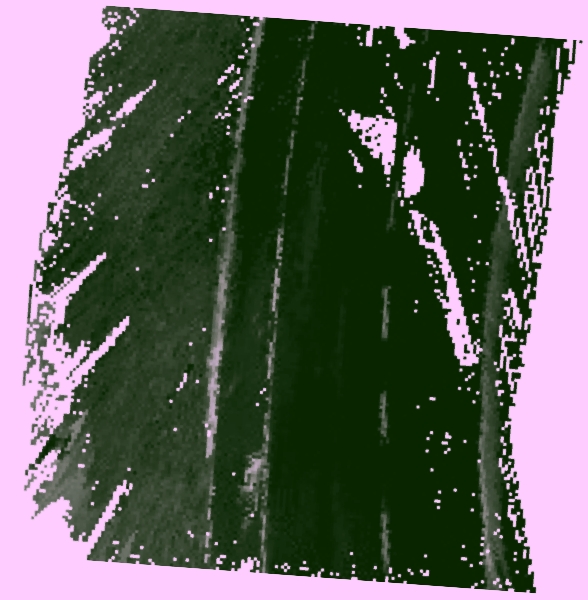}
\caption{An segmented representation of Fig.3's local orthomosaic map, with red as the invalid area and dark green as the valid one.}
\label{fig4} 
\end{figure}

Regardless of whether it is SSD or SAD, the idea of the two is unified, that is, the sum of the difference between each corresponding template pixel and the original pixel in the sliding window W. Unfortunately, this kind of "sum of difference" algorithm is invalid for the problem discussed in this question. To explain this intuitively, we reproduced Fig. \ref{fig3} into two segmented part shown in Fig. \ref{fig4} where red part represents invalid area such as noise and hollow and green for valid. Obviously, the red parts are to neglect. It is now clear that with drastic difference in appearance and style of global(Fig. 2) and local(Fig. 3) maps, there will be tremendous error using SSD or SAD template matching for they will match as more hollow area of the two image as possible for the least sum of differences.

Utilizing the fact that using SSD and SAD algorithms can not effectively solve the map matching problem, we need to apply a higher order cross correlation template matching algorithm.

Unlike SSD and SAD, NCC adopts the sum of dot product correlation operation on the pixels of both template and source image in the sliding window, and normalizes the NCC value with the modulus length as the denominator. Such a correlation method has obtained relatively reliable results for matching two images with huge differences in picture quality but similar characteristics.

\subsection{Weighted Normalized Cross Correlation Algorithm}

Due to the massive contamination of noise and hollows at the edges of each local orthomosaic map constructed by the ground mobile robot, the useless or even harmful cross correlation value calculated there should be weaken at a proper level. On the contrary, information near the center of every local map such as road line or well covers is much more valuable and should be strengthen as key features of the matching. Therefore, the strategy is to increase the weight of the the inner pixels and decrease those outer ones.

Assuming there is a template local map $P$ with the size of $M*N$, its pixel value at $Point(s,t)$ is $P(s,t)$. Let the global map with the size of $W*H$ be $R$ and the sliding window $W_{u,v}$ of each iteration represented by its upper-left pixel $(u,v)$. Therefore, with every pixel in sliding window $W_{u,v}$,for example $W_{u,v}(s,t)$, its corresponding global map pixel should be $R(u+s,v+t)$. Consequently, the searching range for one template should be:
$$
S = (W-M)\cdot(H-N) \eqno{(1)}.
$$

With $\overline{P}$ and $\overline{R}(u,v)$ be the mean value of template local map and the mean value of the corresponding global map of the sliding window $W_{u,v}$ respectively, their declarations are as follows:
$$
\overline{P} = \frac{1}{M\cdot N} \sum\limits_{s=1}^M\sum\limits_{t=1}^NP(s,t) \eqno{(2)}
$$

$$
\overline{R}(u,v) = \frac{1}{M\cdot N} \sum\limits_{s=1}^M\sum\limits_{t=1}^NR(u+s,v+t) \eqno{(3)}.
$$

For the design of weights, this approach adopts the method of calculating the second-order fitting weights with a peak value of 1 and distributed in the second-order interval $(0,1)$. The smaller the Euclidean distance of a pixel is to the center $(\frac{M}{2},\frac{M}{2})$, the greater weight the pixel has. The elaboration of the weighted equation is as follows:
$$
weight_{s,t} = \left|2-\left|\frac{M}{2}-s\right|\cdot\left|\frac{N}{2}-t\right|\right| \eqno{(4)}.
$$

Finally, the equation of the weighted cross correlation(WNCC) is shown as follows:
\begin{small}
$$
\frac{
\sum\limits_{s=1}^M\sum\limits_{t=1}^N
\left|2-\left|\frac{M}{2}-s\right|\cdot\left|\frac{N}{2}-t\right|\right|
\cdot
\left|R(u+s,v+t)-\overline{R}(u,v)\right|
\cdot
\left|p(s,t)-\overline{p}\right|
}{
\sqrt{
\sum\limits_{s=1}^M\sum\limits_{t=1}^N
\left|R(u+s,v+t)-\overline{R}(u,v)\right|
^2
\cdot
\sum\limits_{s=1}^M\sum\limits_{t=1}^N
\left|P(s,t)-\overline{P}\right|
^2
}
} \eqno{(5)}.
$$
\end{small}

In order to realize the real-time matching and localization, GPU based acceleration is implemented, and the result of the acceleration is shown in Section V.

\section{PARTICLE FILTER}

WNCC's precise matching requires traversing all pixels in the image area and calculating cross-correlation values. For robot localization systems with high real-time requirements, its practical application is very limited. Luckily ground mobile robot is able to estimate its localization roughly either by vision odometer or sensors such as IMU. With a rough localization given as a prior location, the searching area for the WNCC could considerably decreased. Therefore, the approach proposed combines particle filter with WNCC to achieve the real-time localization. 

\begin{figure}[t]
\centering
\includegraphics[width=0.5\textwidth]{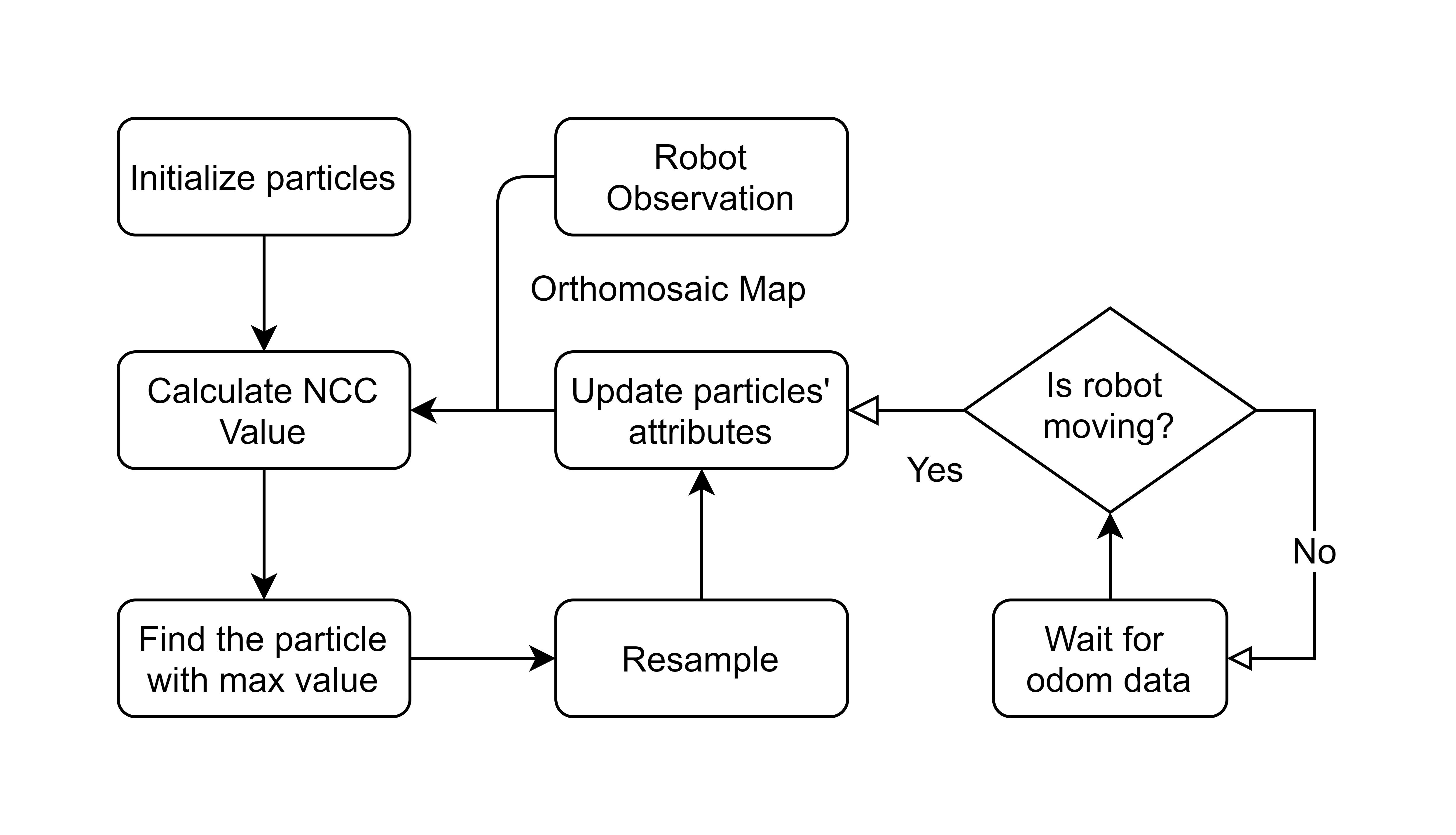}
\caption{The work flow of the particle filter }
\label{fig5} 
\end{figure}

As the workflow shown in Fig. \ref{fig5}, the particle filter localization in this paper integrates the local orthomosaic map as a robot observation into the traditional Monte Carlo localization method\cite{c13}, and use the motion of the ground mobile robot to converge to its correct location. The particle swarm $U=\{p_1, p_2, p_3, ... p_N\}$ represents the confidence of the location and direction of the ground mobile robot in the global orthomosaic map. Considering the prior information about the initial location of the robot, the particles are randomly initialized in a square area near the initial location. The movement of the ground mobile robot is obtained by the odometer while the robot moves according to the odometer information superimposed with the error. After moving to the next location, it makes an observation of the current location to generate a local orthomosaic map. All the particles in the particle swarm are brought into the robot's motion equation one by one to get the next position of the particle swarm. With every particle swarm generated, it is now able to calculate the similarity of the current local orthomosaic map and the sub-global orthomosaic corresponding to each pixel with the proposed approach of WNCC. The resulting similarity of each pixel is then set as the confidence of it. Finally, we proceed an normalization.

In order to prevent degradation during the movement, this article follows the classic roulette method resampling, sampling $M$ times from the particle swarm $U$ to generate a new particle swarm, and the probability of sampling each particle is determined by its weight. By doing this, the particle swarm called the posterior particle swarm will gradually gather to a location with higher similarity. The newly generated particle swarm then moves based on the ground mobile robot's odometer, calculates similarity, and resamples. With this reciprocation, the particles finally converge to the position of the robot with the highest confidence.

To elaborate upon the resampling, some normally distributed position noise $\sigma_{pos}=0.1m$ and rotation noise $\sigma_{rot}=2^\circ$ will be superimposed on the state of particles during resampling to increase the robustness of the system.

\section{RESULT}
In this paper, the proposed WNCC and localization approach combined with particle filter have been verified under experiments of the Aero-Ground Dataset.

\subsection{WNCC}
 A typical example of the Aero-Ground Dataset is used to illustrate the problems when applying SSD, SAD, and the effects of NCC and WNCC are compared on this experimental data.
 
 As stated in Section III, simple template matching such as SSD or SAD that uses sum of difference method will match as more hollow area of the two image as possible for the least sum of differences, as shown in Fig. \ref{fig6}.
 
\begin{figure} 
    \centering
  \subfloat[Result of SSD\label{SSD}]{%
       \includegraphics[width=0.45\linewidth]{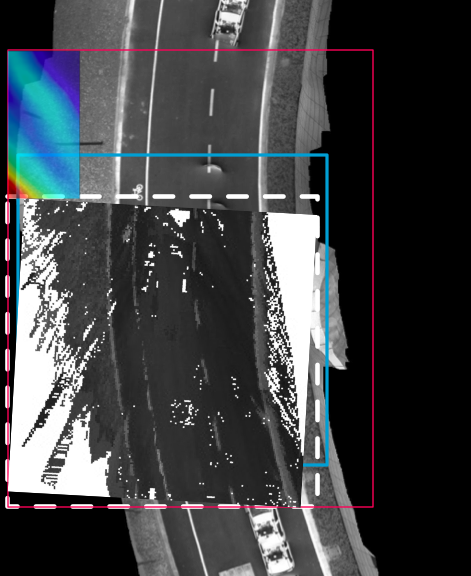}}
    \hfill
  \subfloat[Result of SAD\label{SSD}]{%
        \includegraphics[width=0.45\linewidth]{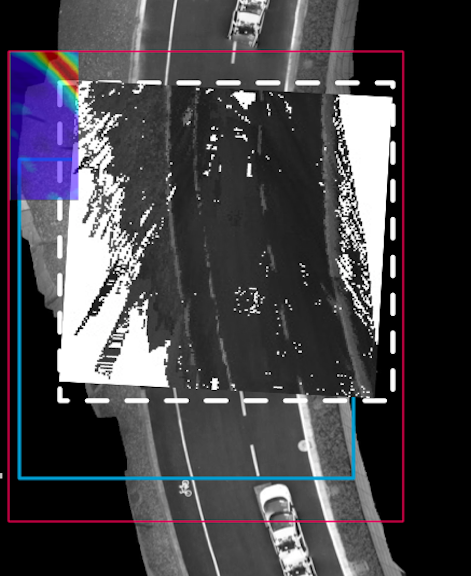}}
  \caption{The result of implementing traditional template matching method SSD and SAD where the red bounding box stands for the regional searching area,blue bounding box stands for the ground truth and the white one stands for the detected matching location of by template matching algorithm.}
  \label{fig6} 
\end{figure}

 The heat map on each of the image respectively shows scores calculate for each location of the sliding window by each method. The deeper the red is, the score of the location is better. Based on the heat map shown from Fig. \ref{fig6}(a) and \ref{fig6}(b), the detected location of the matching unitedly chose places that matched more noise and hollows.
 
 When implementing algorithm NCC, the result becomes better with reasonable location matched as shown in Fig. 7(a). It is clear that the problem with NCC is that it mistakenly matched the curb in the local orthomosaic map with side line of road in the global map. With WNCC implemented, the error is corrected as shown in Fig. 7(b).
 
 \begin{figure}[t]
    \centering
  \subfloat[Result of NCC\label{NCC}]{%
       \includegraphics[width=0.45\linewidth]{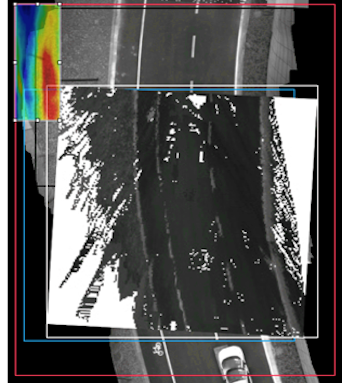}}
    \hfill
  \subfloat[Result of WNCC\label{WNCC}]{%
        \includegraphics[width=0.45\linewidth]{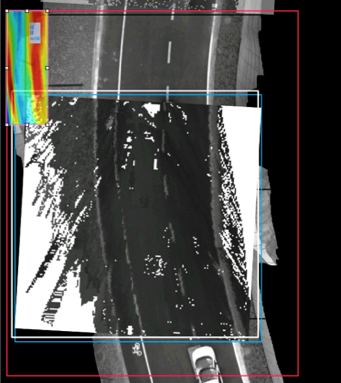}}
  \caption{The result of implementing NCC template matching method and the proposed WNCC template matching method which have the same bounding box as explained in Fig. 6.}
  \label{fig6} 
\end{figure}

By calculating the WNCC algorithm with CUDA, our method is able to speed up the calculation by 500 times, the comparison is shown in TABLE I.

\begin{table}[h]
\caption{The speed test of multiple algorithms tested under the hardware of CPU: AMD RYZEN 3700X, GPU: NVIDIA RTX 2060 Super, RAM: Kingston HyperX DDR4 16G}
\label{speedTable}
\begin{tabular}{|c|c|c|c|c|c|}
\hline
Algorithm & SAD & SSD & NCC & WNCC & \textbf{WNCC with CUDA}  
\\ 
\hline
Time & 812s & 889s &  2996s & 3086s & \textbf{6s} \\ \hline
\end{tabular}
\end{table}

\subsection{Particle Filter Localization}

In the experiment for PF localization with Aero-Ground Dataset, the initial particle swarm number of particle filter is 1000, the position noise $\sigma_{pos}=0.1m$ and the $\sigma_{rot}=2^\circ$. Each pixel on the map has an absolute scale of 10cm corresponds to one pixel. The particle swarm distribution during particle filter localization is shown in Fig. 8.

\begin{figure}[t]
\centering
\includegraphics[width=0.5\textwidth]{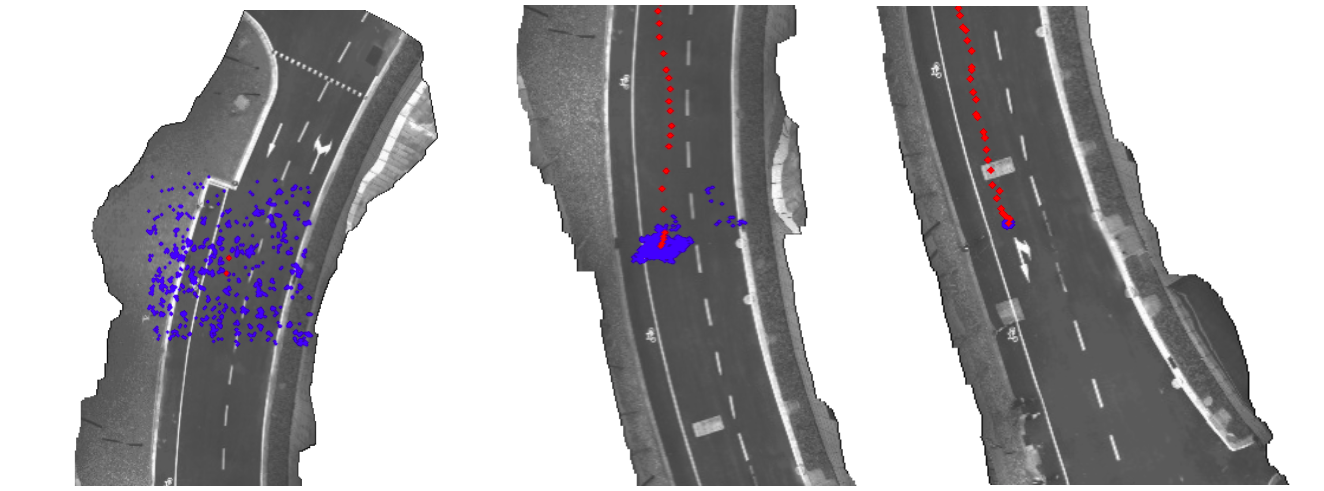}
\caption{Result of the particle swarm distribution and gathering. The blue dots represent particle swarm and red dots represent location of each local map frame. It is proved that with the ground mobile robot moving forward(shown as the order of left to right), the particle swarm gather gradually.}
\label{PGSwarm} 
\end{figure}

The final result of localizing the ground mobile robot within the global map is shown in Fig. 9.
\begin{figure}[h]
\centering
\includegraphics[width=0.5\textwidth]{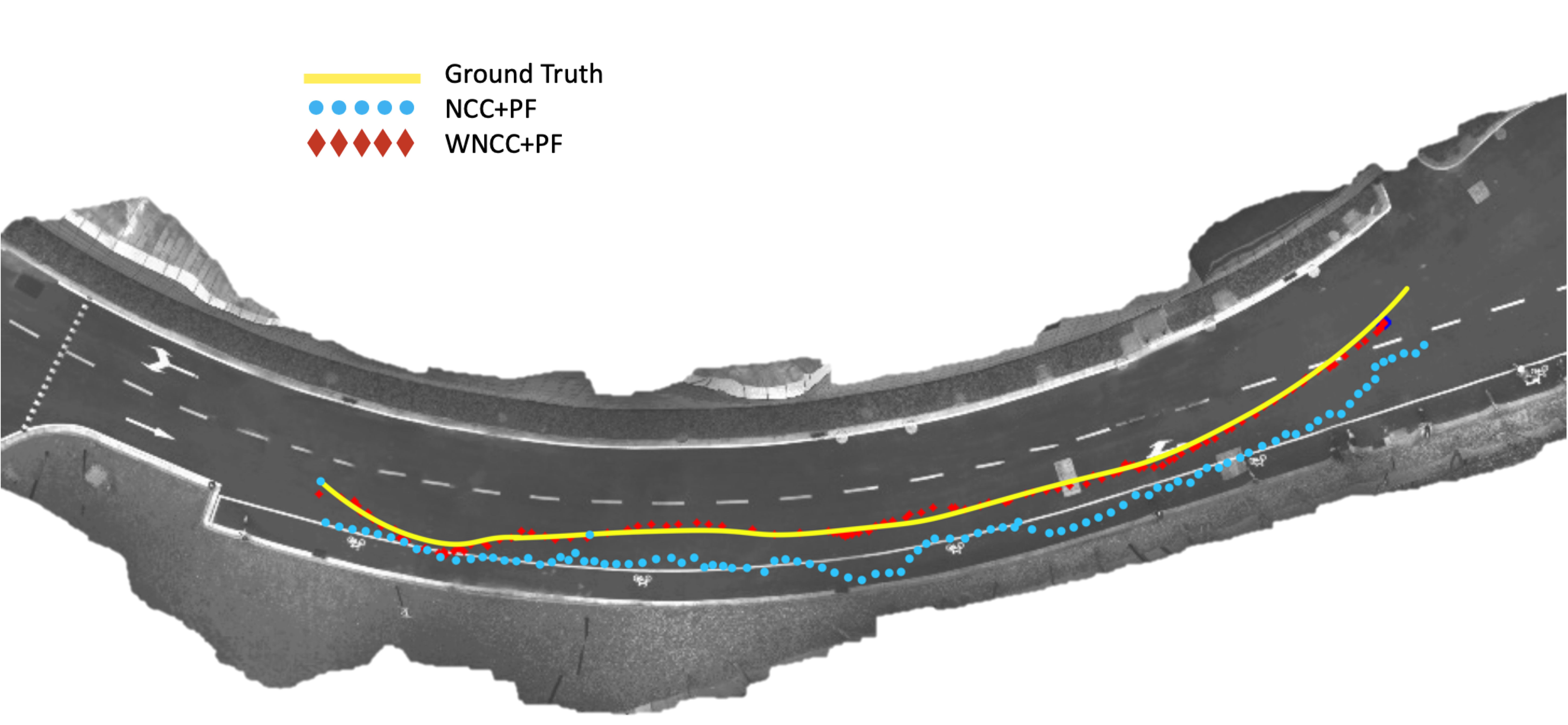}
\caption{The final result of the complete proposed approach restoring the trajectory ground mobile robot within the global map. The comparison with ground truth(yellow line), NCC result(blue dots) and WNCC+PF(red dots) is shown above.}
\label{PGSwarm} 
\end{figure}

The Root Mean Square Error (RMSE) for the trajectory reproduction of the proposed approach of WNCC+PF localization and the comparison of NCC+PF is shown in TABLE II. It is shown that the WNCC method gives an accurate localization with an RMSE of 0.4256 meter and and the comparison method of NCC outcome with an 1.2988 meter.

\begin{table}[h]
\centering
\caption{RMSE of the trajectories reproduced by different approaches}
\label{speedTable}
\begin{tabular}{|c|c|c|}
\hline
Algorithm & \textbf{WNCC+PF} & NCC+PF  
\\ 
\hline
RMSE & \textbf{0.4256m} & 1.2988m \\ \hline
\end{tabular}
\end{table}

\addtolength{\textheight}{-13cm}   

\section{CONCLUSIONS}
This paper proposes and validates an aerial-ground collaboration system based on orthomosaic map matching and particle filter localization. Through this system, the ground mobile robot can locate itself within the global map constructed by the MAV's bird's-eye view in real time. This system is very lightweight and can complete high-precision localization without the need for costly equipments such as structured light and lidar.

This system has good time robustness and scalability. When the lighting conditions permit, this system can rely on the vision imaging system to complete the task. When it is poor-lit, the system also supports Lidar together with a texture-use only camera to complete the task. This system also has good spatial robustness, which means that the system does not limit the flying height, flight distance, and flying place of the MAV. It supports both indoor and outdoor aerial-ground collaboration with a fast response and accurate localization.



\begin{thebibliography}{13}

\bibitem{c1} https://github.com/ZJU-Robotics-Lab/OpenDataSet
\bibitem{c2} E. Mueggler, M. Faessler, F. Fontana, and D. Scaramuzza, “Aerial- guided navigation of a ground robot among movable obstacles,” IEEE International Symposium on Safety, Security, and Rescue Robotics (SSRR), pp. 1–8, 2014.
\bibitem{c3} P. Rudol, M. Wzorek, G. Conte, and P. Doherty, “Micro Unmanned Aerial Vehicle Visual Servoing for Cooperative Indoor Exploration,” IEEE Aerospace Conference, 2008.
\bibitem{c4} Fankhauser, Péter, et al. "Collaborative navigation for flying and walking robots." 2016 IEEE/RSJ International Conference on Intelligent Robots and Systems (IROS). IEEE, 2016..
\bibitem{c5} N.Michael,S.Shen,K.Mohta,Y.Mulgaonkar,V.Kumar,K.Nagatani, Y. Okada, S. Kiribayashi, K. Otake, K. Yoshida, K. Ohno, E. Takeuchi, and S. Tadokoro, “Collaborative mapping of an earthquake-damaged building via ground and aerial robots,” Journal of Field Robotics, vol. 29, no. 5, pp. 832–841, 2012.
\bibitem{c6} Käslin, Roman, et al. "Collaborative localization of aerial and ground robots through elevation maps." 2016 IEEE International Symposium on Safety, Security, and Rescue Robotics (SSRR). IEEE, 2016.
\bibitem{c7} P. Fankhauser and M. Hutter, "A Universal Grid Map Library: Implementation and Use Case for Rough Terrain Navigation", in Robot Operating System (ROS) – The Complete Reference (Volume 1), A. Koubaa (Ed.), Springer, 2016.
\bibitem{c8} A. A. Goshtasby, “Similarity and Dissimilarity Measures,” in Image Registration: Principles, Tools and Methods, A. A. Goshtasby, Ed. London: Springer London, 2012, pp. 7–66.
\bibitem{c9} C. Forster, M. Pizzoli, and D. Scaramuzza, “Air-ground localiza- tion and map augmentation using monocular dense reconstruction,” IEEE/RSJ International Conference on Intelligent Robots and Systems (IROS), 2013.
\bibitem{c10} A. Wendel, A. Irschara, and H. Bischof, “Automatic alignment of 3d reconstructions using a Digital Surface Model,” in IEEE Computer Society Conference on Computer Vision and Pattern Recognition Workshops, 2011.
\bibitem{c11} Metashape. 1.6.0 build 9925. 30 days free trial. January 2020. Agisoft. https://www.agisoft.com/. 01/02/2020.
\bibitem{c12} Y.Pan, X. Xu, Y. Wang, X. Ding, and R. Xiong, “Gpu accelerated real-time traversability mapping,” in 2019 IEEE International Conference on Robotics and Biomimetics (ROBIO), pp. 734-740, IEEE, 2019.
\bibitem{c13} S. Thrun, D. Fox, W. Burgard, and F. Dellaert, “Robust Monte Carlo localization for mobile robots,” Artificial Intelligence, vol. 128, no. 1, pp. 99–141, 2001.



\end{thebibliography}
\end{document}